\documentclass[10pt,twocolumn,letterpaper]{article}

\usepackage{iccv}
\usepackage{times}
\usepackage{epsfig}
\usepackage{graphicx}
\usepackage{amsmath}
\usepackage{amssymb}

\usepackage[breaklinks=true,bookmarks=false]{hyperref}

\iccvfinalcopy 

\def\iccvPaperID{****} 

\ificcvfinal\fi

\begin{document}

\title{Technical Report for ICCV 2021 Challenge SSLAD-Track3B: \\Transformers Are Better Continual Learners}

\author{
Duo Li\thanks{These authors contribute equally to this work.} , Guimei Cao\footnotemark[1] , Yunlu Xu , Zhanzhan Cheng\thanks{Contact author.}, Yi Niu\\
AML Group, Hikvision Research Institute\\
Shanghai, China\\
{\tt\small \{liduo6, caoguimei, xuyunlu, chengzhanzhan, niuyi\}@hikvision.com}
}
\maketitle
\ificcvfinal\thispagestyle{empty}\fi

\begin{abstract}
  In the SSLAD-Track 3B challenge on continual learning, we propose the method of \textbf{CO}ntinual \textbf{L}earning with \textbf{T}ransformer (COLT).
  We find that transformers suffer less from catastrophic forgetting compared to convolutional neural network.
  The major principle of our method is to equip the transformer based feature extractor with old knowledge distillation and head expanding strategies to compete catastrophic forgetting.
  In this report, we first introduce the overall framework of continual learning for object detection.
  Then, we analyse the key elements' effect on withstanding catastrophic forgetting in our solution.
  Our method achieves 70.78 mAP on the SSLAD-Track 3B challenge test set.
\end{abstract}

\section{Introduction}

Catastrophic forgetting is one of the major differences between artificial neural networks and human brains \cite{van2020brain}.
To overcome catastrophic forgetting in artificial neural networks, three types of methods have been explored in the past few years, replay \cite{rebuffi2017icarl, bang2021rainbow, cong2020gan, van2020brain, liu2020generative}, regularization \cite{li2017learning, tao2020topology, hu2021distilling, castro2018end, shi2021continual, lopez2017gradient, chaudhry2018efficient, farajtabar2020orthogonal, saha2020gradient}, and expand \cite{yoon2018lifelong, mallya2018packnet, abati2020conditional, yan2021dynamically}.
All of the above methods try to answer the question, `what kind of knowledge storage, what kind of training strategies, and what kind of network structures are suitable to keep old memories for neural networks?'
In this report, our answer is to use the transformer as the feature extractor, conduct knowledge distillation on old samples, and reduce the domain gap by adaptively expanding.

The SSLAD-Track 3B challenge at ICCV 2021 requires to design continual learning algorithms for object detection in automatic driving scenarios.
There are in total four different scenarios.
The model shall learn from each scenario one by one. In the end, the model is evaluated on the four test sets for each scenario.
There are totally 7.8k images in the training set of the four scenarios.
In each scenario, there are six classes of objects, pedestrian, cyclist, car, truck, tram, and tricycle.
In the challenge, mean AP across all four test set of the four scenarios is used as the indicator.

\section{Methods}

\begin{figure*}[t]
\begin{center}
    \includegraphics[width=1.0\linewidth]{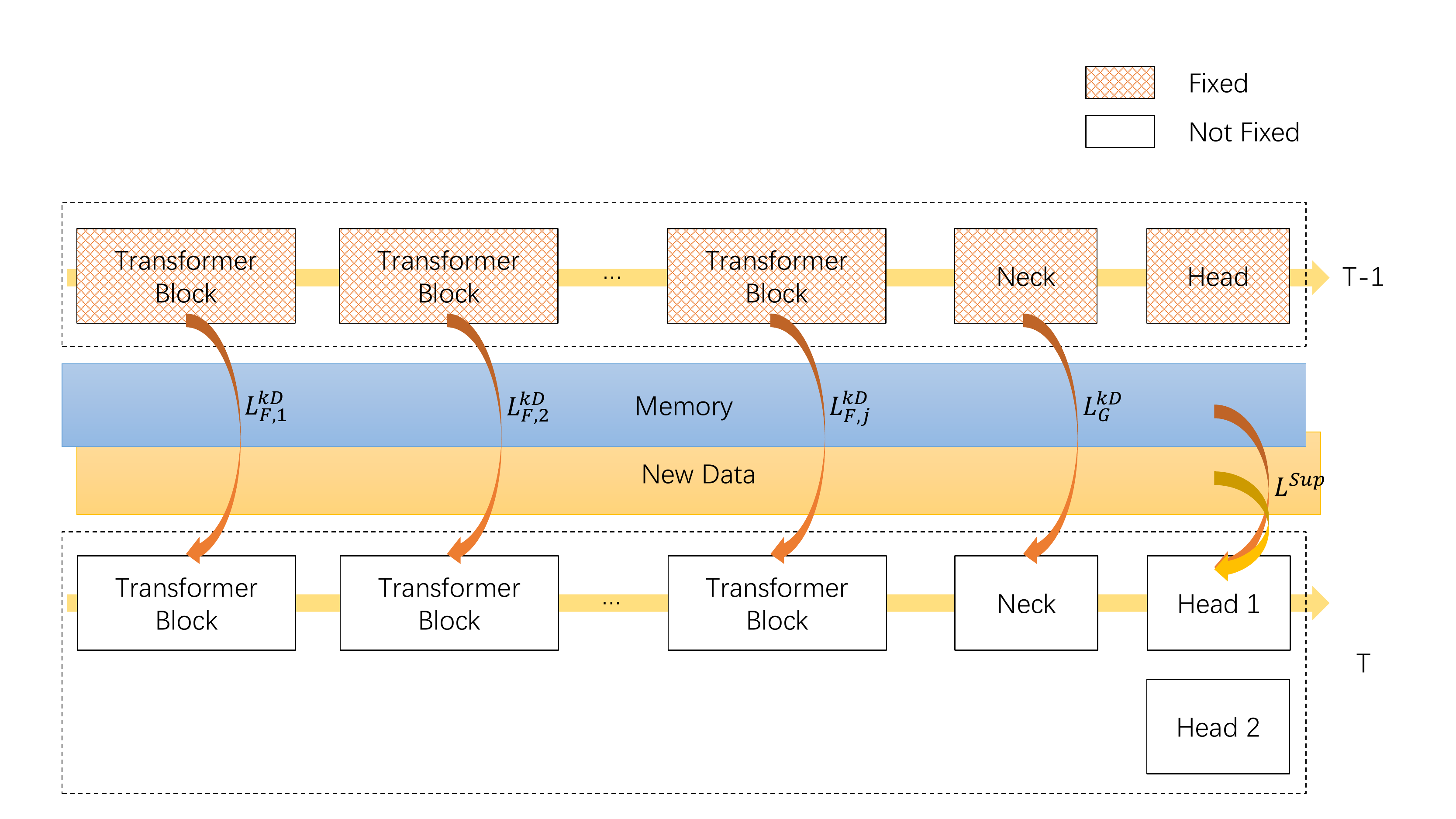}
\end{center}
   \caption{Overall framework of COLT.}
\label{fig:framework}
\end{figure*}

In this section, we present our method to deal with catastrophic forgetting in object detection.
As shown in Fig. \ref{fig:framework}, there are three major components in our framework.
As we observe that transformers suffer less from the forgetting problem, we use the transformer as the feature extractor.
Following regularization based continual learning methods, we employ sample replay as well as old knowledge distillation strategies.
To reduce the domain gap among scenarios, we adaptively expand the detector heads according to the current model's validation loss on the training set of the new scenario.

We follow the rules of the SSLAD-Track 3B challenge, the network structure only changes when the model adaptively expand new heads when large domain gaps are detected, and we only store 250 images in the memory through the whole continual training period.

\subsection{Transformer: Less Forgetting Feature Extractor}

Using convolutional neural network (CNN) as the feature extractor is common in continual learning methods for computer vision tasks.
However, two characteristics of CNN limits its performance in continual learning.
First, large CNN models easily get over-fitted on a domain when there lack a large mount of training data.
One way to compete for catastrophic forgetting in continual learning is fixing the model's backbone during training and only finetune the detector's neck and head.
This brings a side effect that the model gets bad performance on the new scenario.
It can be inferred that if the backbone can extract features which generalize well to unseen domains, then there is little need to adjust the backbone to fit new scenarios in continual learning. And the model still gets good performance on new scenarios.
Second, another reason causes the forgetting problem in the convolutional neural network is the batchnorm (BN) layer.
It has been observed that if we fix all the BN layers in continual training, the forgetting problem gets obviously remitted.

Compared to CNN, transformer has been shown to have a good ability to generalize well to new domains, and it does not have BN layers.
Since it naturally overcomes the two elements for forgetting, we use the Swin Transformer \cite{liu2021Swin} as the backbone for CascadeRCNN.

\subsection{Adaptive Head Expanding}

Other than catastrophic forgetting, there exists another problem in continual learning, which is the problem of domain gap.
Although domain adaptation methods can partly solve this problem, it is still face with a trade-off between two domains.
Inspired by the fact that expert models usually perform better than one single model \cite{2016Expert}, we implement multiple heads for different domains whose gap is large.
During training, we first estimate the domain gap between the new task and the old tasks. The estimation is conducted by calculating the average validation loss of the current model on the new task. If domain gap is large, than the model expands a new head and this head specifically learns to predict the samples of this domain. And the original head learns to keep old knowledge on the old tasks.
During testing time, given the sample's task-id, the model chooses which head to predict the results.

\subsection{Old Knowledge Distillation}

Following the work in \cite{li2017learning, tao2020topology, hu2021distilling, castro2018end, shi2021continual}, knowledge distillation on old samples effectively competes catastrophic forgetting.
For simplicity, we only conduct knowledge distillation (KD) on the backbone and neck of the detector.
In the training period of each scenario, the model before training is copied and fixed to be used as the teacher model.
The other model (student) is trained on the new scenario, together with the KD loss as a regularization term.
The KD loss is defined as follows:
\begin{equation}
  L_{i}^{KD} = \sum_{j=1}^{M}{||F^t_j(x_i)-F^s_j(x_i)||^2} + \sum_{k=1}^{N}{||G^t_j(x_i)-G^s_j(x_i)||^2}
\end{equation}
where $x_i$ is a sample from old scenarios which is stored in the rehearsal memory.
$F^t$ and $G^t$ are backbone and neck features extracted by the teacher model, and $F^s$ and $G^s$ are backbone and neck features extracted by the student model.

In above definition, the first term indicates the backbone KD loss, and the second term indicates the neck KD loss.
The memory size is limited to 250 samples, according to the rules of the competition.

\begin{table*}
\caption{Ablation study on network structure and other strategies. The models are evaluated on the validation set.}
\begin{center}
\begin{tabular}{l||c|c|c|c||c|c}
\hline
Method & Detector & Backbone & KD & Head Expanding & Mean AP $\uparrow$ & FR $\downarrow$ \\
\hline\hline
1 & Yolov3 & Darknet19 & w/o & w/o & 44.28 & 3.91 \\
\hline
2 & FasterRCNN & ResNet50 & w/o & w/o & 54.80 & 4.97 \\
3 & FasterRCNN & ResNet101 & w/o & w/o & 54.22 & 17.76 \\
4 & FasterRCNN & ResNet101 & w & w/o & 57.95 & 14.58 \\
5 & CascadeRCNN & ResNet101 & w/o & w/o & 54.28 & 17.45 \\
6 & CascadeRCNN & ResNet101 & w & w/o & 58.00 & 14.32 \\
\hline
7 & CascadeRCNN & Transformer & w/o & w/o & 72.67 & 2.94 \\
8 & CascadeRCNN & Transformer & w & w/o & 72.99 & 1.65 \\
9 & CascadeRCNN & Transformer & w/o & w & 73.59 & 1.71 \\
COLT & CascadeRCNN & Transformer & w & w & \textbf{75.11} & \textbf{0.46} \\
\hline
\end{tabular}
\end{center}
\end{table*}

\section{Experiments}

\subsection{Settings}
In the SSLAD-Track 3B challenge, SODA10M is a 2D object detection dataset, which contains images captured from four different scenarios.
These scenarios are set as four tasks for continual learning. The continual learner is trained on each task sequentially.
The tasks are:
\begin{itemize}
  \item Task 1: Daytime, citystreet and clear weather. There are 4470 images in the training set.
  \item Task 2: Daytime, highway and clear/overcast weather. There are 1329 images in the training set.
  \item Tast 3: Night. There are 1479 images in the training set.
  \item Tast 4: Daytime, rain. There are 524 images in the training set.
\end{itemize}

After training on all four tasks, the model is evaluated on the validation set and test set. Mean AP over the four tasks is used as the final indicator.
In this report, we also introduce forgetting rate (FR) to compare the relative forgetting degree between different methods.
Forgetting rate is defined as follows:
\begin{equation}\label{eq}
  FR = \frac{1}{T-1}\sum_{i=1}^{T-1}\frac{D_i(M_i)-D_i(M_{T})}{D_i(M_i)}
\end{equation}
Forgetting rate (FR) indicates the disparity between current model's performance and its historical performance.
If FR is large, it means that current model suffers a lot from catastrophic forgetting, and falls far behind the upper bound of itself.
If FR is small, it means that current model reaches the upper bound performance which it shall have.

\subsection{Implementation Details}
The algorithm is implemented using Avalanche \cite{lomonaco2021avalanche}.
We use the CascadeRCNN as the detector in continual learning.
For the backbone, we use the Swin Transformer \cite{liu2021Swin}.
To improve the generalization ability of the transformer, we first pre-train it on ImageNet.
During pre-training, we conduct data augmentations like multi-scale, flip, color jittering, and MixUp on instances.
For continual learning on the SSLAD dataset, we train the model on 8 GPUs, one sample for each GPU.
During training each scenario, the learning rate is set to 0.001 and divide by 10 after 33 and 44 epochs.
Training is stopped at epoch 50.
The ratio between the supervised learning loss and knowledge distillation loss is set to 1:20.
When adaptively expanding the detector heads, the threshold of the average validation loss is set to 1.2.

\subsection{Results}

We report the comparative results on the SSLAD-Track3B validation and test sets.
Our method (COLT) gets 75.11 mean AP on the validation set. As a comparison, the baseline provided on the challenge web-site achieves 55.53 mean AP.
We get 19.58 mean AP higher than the baseline method on the validation set.
On the test set, we get 70.78 mean AP.
The detailed results on the test set are shown in Table 2.

\begin{table*}
\label{testset}
\caption{Results on the SSLAD-Track3B test set.}
\begin{center}
\begin{tabular}{l||c||c|c|c|c|c|c|c|c|c|c|}
\hline
Method & Mean AP & Task1 & Task2 & Task3 & Task4 & pedestrian& cyclist & car & truck & tram & tricycle \\
\hline\hline
COLT & 70.78 & 77.24 & 68.12 & 68.52 & 69.24 & 73.79 & 78.20 & 88.37 & 77.78 & 71.50 & 35.04 \\
\hline
\end{tabular}
\end{center}
\end{table*}

\subsection{Ablation Study}

\textbf{Model size.}
To study the influence of the model size on continual learning, we compare the performance between Faster RCNN with ResNet50 and Faster RCNN with ResNet101.
Results in Table 1 (method 2 v.s. 3) show that larger model does not guarantee better continual learning performance.
We can see that the large model (17.76) even performs worse than the small model (4.97) on FR (smaller is better).
An explanation is that when the model becomes larger, it not only gets higher model capacity, but also becomes easier over-fitted to current scenario.
Over-fitting leads to good performance on current task, while causes the forgetting problem of old tasks.

As comparison, we train an even larger detector using CascadeRCNN with transformer as backbone.
In Table 1 (method 5 v.s. 7), when we switch from ResNet101 (small) to transformer (large), both mean AP and FR indicate that the latter is better.
Through these experiments, we conclude that the major element that influences the model's forgetting problem is not model size, but the generalization ability of the feature extractor.


\textbf{Head Expanding.}
Since head expanding requires the task-id at test time, it is an optional module in our framework.
In Table 1 (method 7 v.s. 9, 8 v.s. COLT), head expanding is shown to be effective.
The major improvement is on task 3.
It is because task 3 is the scenario of the night, which is largely different from the other tasks.
So the model adaptively expand a new head for task 3 before training on the night scenario.

\textbf{Old Knowledge Distillation.}
Previous work show that knowledge distillation on CNN models is an efficient way to compete for catastrophic forgetting.
Our experiment in Table 1 (method 7 v.s. 8, 9 v.s. COLT) shows that it still holds for transformer based models.
The knowledge distillation strategy improves the mean AP from 73.59 to 75.11, and FR from 1.71 to 0.46.

\section{Conclusion}
In this report, we present our continual learning object detection method on SSLAD-Track 3B challenge.
Our method consists of three major components: transformer based feature extractor, old knowledge distillation, and adaptively growing multiple heads architecture.
Through experiments, we show that transformers suffer less from catastrophic forgetting.
Our method achieves 70.78 mean AP on the SSLAD-Track 3B challenge test set,
which helps us get the 1st place in this challenge.

\textbf{Future work.}
In this report, we try to emphasis the principle of the transformer in overcoming catastrophic forgetting, but we still know little about the transformer's characteristics in continual learning.
One future direction is to study the transformer inspired network which is specifically designed for continual learning.
During this challenge, we also tried to generate pseudo samples with VAEs and GANs, which have been proven effective in previous work \cite{cong2020gan, shin2017continual, liu2020generative, van2020brain} on continual learning for classification task.
However, we suffered from problems of variant of instance's aspect ratio, image generation for object detection, long-tail problem for generative models, and so on.
Generative sample/feature replay for object detection is also a critical problem which needs to be studied in the future.


{\small
\bibliographystyle{ieee_fullname}
\bibliography{egbib}

\begin{thebibliography}{10}\itemsep=-1pt

\bibitem{abati2020conditional}
Davide Abati, Jakub Tomczak, Tijmen Blankevoort, Simone Calderara, Rita
  Cucchiara, and Babak~Ehteshami Bejnordi.
\newblock Conditional channel gated networks for task-aware continual learning.
\newblock In {\em Proceedings of the IEEE Conference on Computer Vision and
  Pattern Recognition}, 2020.

\bibitem{2016Expert}
R. Aljundi, P. Chakravarty, and T. Tuytelaars.
\newblock Expert gate: Lifelong learning with a network of experts.
\newblock In {\em Proceedings of the IEEE Conference on Computer Vision and
  Pattern Recognition}, 2016.

\bibitem{bang2021rainbow}
Jihwan Bang, Heesu Kim, YoungJoon Yoo, Jung-Woo Ha, and Jonghyun Choi.
\newblock Rainbow memory: Continual learning with a memory of diverse samples.
\newblock In {\em Proceedings of the IEEE Conference on Computer Vision and
  Pattern Recognition}, 2021.

\bibitem{castro2018end}
Francisco~M Castro, Manuel~J Mar{\'\i}n-Jim{\'e}nez, Nicol{\'a}s Guil, Cordelia
  Schmid, and Karteek Alahari.
\newblock End-to-end incremental learning.
\newblock In {\em Proceedings of the European Conference on Computer Vision},
  2018.

\bibitem{chaudhry2018efficient}
Arslan Chaudhry, Marc'~Aurelio Ranzato, Marcus Rohrbach, and Mohamed Elhoseiny.
\newblock Efficient lifelong learning with a-gem.
\newblock In {\em International Conference on Learning Representations}, 2018.

\bibitem{cong2020gan}
Yulai Cong, Miaoyun Zhao, Jianqiao Li, Sijia Wang, and Lawrence Carin.
\newblock Gan memory with no forgetting.
\newblock In {\em Neural Information Processing Systems}, 2020.

\bibitem{farajtabar2020orthogonal}
Mehrdad Farajtabar, Navid Azizan, Alex Mott, and Ang Li.
\newblock Orthogonal gradient descent for continual learning.
\newblock In {\em International Conference on Artificial Intelligence and
  Statistics}, 2020.

\bibitem{hu2021distilling}
Xinting Hu, Kaihua Tang, Chunyan Miao, Xian-Sheng Hua, and Hanwang Zhang.
\newblock Distilling causal effect of data in class-incremental learning.
\newblock In {\em Proceedings of the IEEE Conference on Computer Vision and
  Pattern Recognition}, 2021.

\bibitem{li2017learning}
Zhizhong Li and Derek Hoiem.
\newblock Learning without forgetting.
\newblock {\em IEEE Transactions on Pattern Analysis and Machine Intelligence},
  2017.

\bibitem{liu2020generative}
Xialei Liu, Chenshen Wu, Mikel Menta, Luis Herranz, Bogdan Raducanu, Andrew~D
  Bagdanov, Shangling Jui, and Joost~van de Weijer.
\newblock Generative feature replay for class-incremental learning.
\newblock In {\em Proceedings of the IEEE Conference on Computer Vision and
  Pattern Recognition Workshops}, 2020.

\bibitem{liu2021Swin}
Ze Liu, Yutong Lin, Yue Cao, Han Hu, Yixuan Wei, Zheng Zhang, Stephen Lin, and
  Baining Guo.
\newblock Swin transformer: Hierarchical vision transformer using shifted
  windows.
\newblock {\em arXiv preprint arXiv:2103.14030}, 2021.

\bibitem{lomonaco2021avalanche}
Vincenzo Lomonaco, Lorenzo Pellegrini, Andrea Cossu, Antonio Carta, Gabriele
  Graffieti, Tyler~L. Hayes, Matthias~De Lange, Marc Masana, Jary Pomponi, Gido
  van~de Ven, Martin Mundt, Qi She, Keiland Cooper, Jeremy Forest, Eden
  Belouadah, Simone Calderara, German~I. Parisi, Fabio Cuzzolin, Andreas
  Tolias, Simone Scardapane, Luca Antiga, Subutai Amhad, Adrian Popescu,
  Christopher Kanan, Joost van~de Weijer, Tinne Tuytelaars, Davide Bacciu, and
  Davide Maltoni.
\newblock Avalanche: an end-to-end library for continual learning.
\newblock In {\em Proceedings of IEEE Conference on Computer Vision and Pattern
  Recognition}, 2nd Continual Learning in Computer Vision Workshop, 2021.

\bibitem{lopez2017gradient}
David Lopez-Paz and Marc'Aurelio Ranzato.
\newblock Gradient episodic memory for continual learning.
\newblock {\em Advances in neural information processing systems}, 2017.

\bibitem{mallya2018packnet}
Arun Mallya and Svetlana Lazebnik.
\newblock Packnet: Adding multiple tasks to a single network by iterative
  pruning.
\newblock In {\em Proceedings of the IEEE Conference on Computer Vision and
  Pattern Recognition}, 2018.

\bibitem{rebuffi2017icarl}
Sylvestre-Alvise Rebuffi, Alexander Kolesnikov, Georg Sperl, and Christoph~H
  Lampert.
\newblock icarl: Incremental classifier and representation learning.
\newblock In {\em Proceedings of the IEEE conference on Computer Vision and
  Pattern Recognition}, 2017.

\bibitem{saha2020gradient}
Gobinda Saha, Isha Garg, and Kaushik Roy.
\newblock Gradient projection memory for continual learning.
\newblock In {\em International Conference on Learning Representations}, 2021.

\bibitem{shi2021continual}
Yujun Shi, Li Yuan, Yunpeng Chen, and Jiashi Feng.
\newblock Continual learning via bit-level information preserving.
\newblock In {\em Proceedings of the IEEE Conference on Computer Vision and
  Pattern Recognition}, 2021.

\bibitem{shin2017continual}
Hanul Shin, Jung~Kwon Lee, Jaehong Kim, and Jiwon Kim.
\newblock Continual learning with deep generative replay.
\newblock In {\em Neural Information Processing Systems}, 2017.

\bibitem{tao2020topology}
Xiaoyu Tao, Xinyuan Chang, Xiaopeng Hong, Xing Wei, and Yihong Gong.
\newblock Topology-preserving class-incremental learning.
\newblock In {\em Proceedings of the European Conference on Computer Vision}.
  Springer, 2020.

\bibitem{van2020brain}
Gido~M van~de Ven, Hava~T Siegelmann, and Andreas~S Tolias.
\newblock Brain-inspired replay for continual learning with artificial neural
  networks.
\newblock {\em Nature communications}, 2020.

\bibitem{yan2021dynamically}
Shipeng Yan, Jiangwei Xie, and Xuming He.
\newblock Der: Dynamically expandable representation for class incremental
  learning.
\newblock In {\em Proceedings of the IEEE Conference on Computer Vision and
  Pattern Recognition}, 2021.

\bibitem{yoon2018lifelong}
Jaehong Yoon, Eunho Yang, Jeongtae Lee, and Sung~Ju Hwang.
\newblock Lifelong learning with dynamically expandable networks.
\newblock In {\em International Conference on Learning Representations}, 2018.

\end{thebibliography}
}

\end{document}